\documentclass[letterpaper, 10 pt, conference]{ieeeconf}  

\IEEEoverridecommandlockouts                              

\usepackage{graphics} 
\usepackage{amsmath,amssymb,amsfonts} 
\usepackage{bm}
\usepackage{subcaption}	 
\usepackage{graphicx}
\usepackage{multirow}
\usepackage{array}
\usepackage{caption}

\usepackage{url}
\usepackage{hyperref}
\hypersetup{colorlinks,urlcolor=red}
\usepackage{xcolor}
\usepackage{gensymb}
\usepackage{svg}
\svgsetup{
    inkscapepath=i/svg-inkscape/
}
\svgpath{{svg/}}

\usepackage[backend=bibtex,natbib=true,style=numeric-comp,sorting=none,giveninits=true,maxbibnames=99,url=false,doi=false]{biblatex}
\title{\LARGE \bf
RadarCam-Depth: Radar-Camera Fusion for Depth Estimation \\
with Learned Metric Scale
}

\author{Han Li$^{1,\dagger}$, Yukai Ma$^{1,\dagger}$, Yaqing Gu$^{1}$, Kewei Hu$^{1}$, Yong Liu$^{1,*}$, Xingxing Zuo$^{2,*}$
\thanks{ $^{1}$The authors are with the Institute of Cyber-Systems and Control, Zhejiang University, Hangzhou, China. 
}%
\thanks{$^{2}$ The author is with the Department of Computing and Mathematical Sciences,
California Institute of Technology, USA.}%
\thanks{$^*$ Xingxing Zuo and Yong Liu are the corresponding authors (Email: {\tt\small zuox@caltech.edu; yongliu@iipc.zju.edu.cn}).}
\thanks{$^\dagger$ These authors contributed equally to this work.}
\thanks{This work is supported by NSFC 62088101 Autonomous Intelligent Unmanned Systems.}
}

\begin{document}

\maketitle
\thispagestyle{empty}
\pagestyle{empty}

\begin{abstract}

We present a novel approach for metric dense depth estimation based on the fusion of a single-view image and a sparse, noisy Radar point cloud. The direct fusion of heterogeneous Radar and image data, or their encodings, tends to yield dense depth maps with significant artifacts, blurred boundaries, and suboptimal accuracy.
To circumvent this issue, we learn to augment versatile and robust monocular depth prediction with the dense metric scale induced from sparse and noisy Radar data.
We propose a Radar-Camera framework for highly accurate and fine-detailed dense depth estimation with four stages, including monocular depth prediction, global scale alignment of monocular depth with sparse Radar points, quasi-dense scale estimation through learning the association between Radar points and image patches, and local scale refinement of dense depth using a scale map learner.
Our proposed method significantly outperforms the state-of-the-art Radar-Camera depth estimation methods by reducing the mean absolute error (MAE) of depth estimation by 25.6\% and 40.2\% on the challenging nuScenes dataset and our self-collected ZJU-4DRadarCam dataset, respectively. Our code and dataset will be released at \url{https://github.com/MMOCKING/RadarCam-Depth}.

\end{abstract}
\vspace{-3pt}

\section{Introduction}

Perceiving the environment is critically important for autonomous driving, where accurate depth estimation is fundamental for dense reconstruction, 3D detection, and obstacle avoidance. Cameras and range sensors have been widely used for perceiving dense depth.
Learned monocular depth (mono-depth) estimation methods based on CNN networks \cite{teed2018deepv2d, wong2020unsupervised, birkl2023midas, guizilini2023towards, wofk2023monocular} have been prevalent in recent years due to their versatile applicability and plausible accuracy. They benefit from the solid contextual priors from extensive training on diverse datasets.
While mono-depth networks excel in estimating up-to-scale depth, they fail to predict the accurate metric scale of depth. This limit arises from the inherent challenge of capturing scale with single-view cameras and the difficulty of learning the diverse scale in complex scenarios. 

Range sensors like LiDAR and Radar can provide metric scale information of the scene \cite{ma2023rolm, ma2023fmcw}. While LiDAR is renowned for its ability to generate dense and accurate point clouds, its widespread deployment faces challenges due to high costs, power consumption, and data bandwidth limits. In contrast, 3D Radar has witnessed remarkable advancements, making it attractive in autonomous driving, owing to its affordability, low power consumption, and high resilience in challenging fog and smoke scenarios. The emerging 4D Radar additionally provides an elevation dimension with extended applicability.
Fusing data from a single camera and a Radar for metric dense depth estimation becomes a promising research avenue \cite{lin2020depth, long2021full, long2021radar, lo2021depth, singh2023depth, gasperini2021r4dyn}. It holds substantial significance in autonomous driving since its appealing characteristics, like cost-effectiveness,  complementarity in sensing capabilities, and remarkable robustness and reliability.

\begin{figure}[t]
      \centering
      \includegraphics[width=0.45\textwidth]{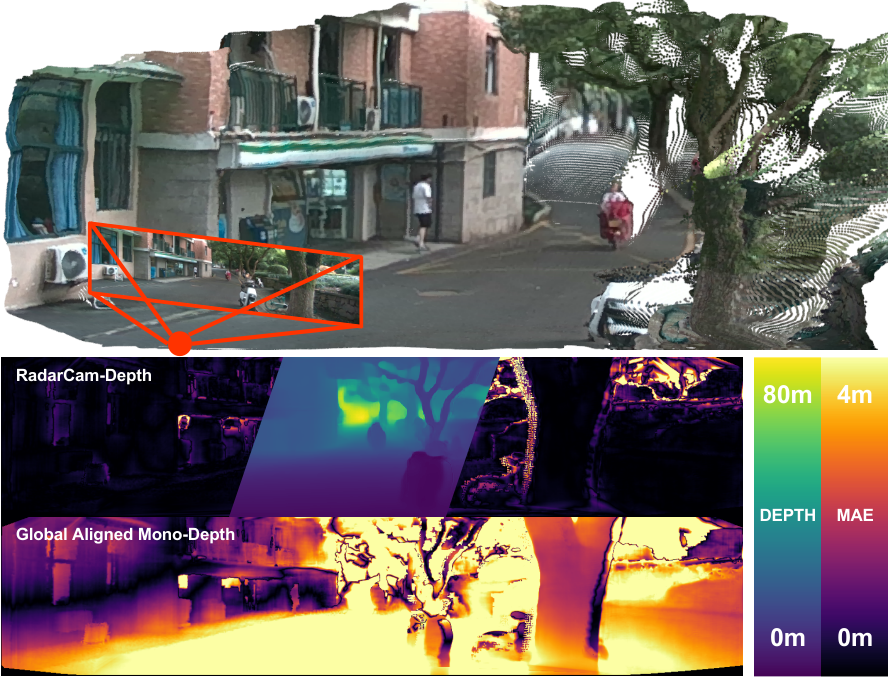}
      \captionsetup{font=small}
      \caption{Top: 3D visualization of the metric depth estimation from our proposed RadarCam-Depth; Middle: Our metric depth estimation overlaid on corresponding error map; Bottom: Error map of Mono-depth after scale-aligned to Radar points. Our depth estimation exhibits exceptional metric accuracy and fine details.}
      \label{fig:fm}
      \vspace{-20pt} 
\end{figure}

\begin{figure*}[t]
      \centering
      \includegraphics[width=0.88\textwidth]{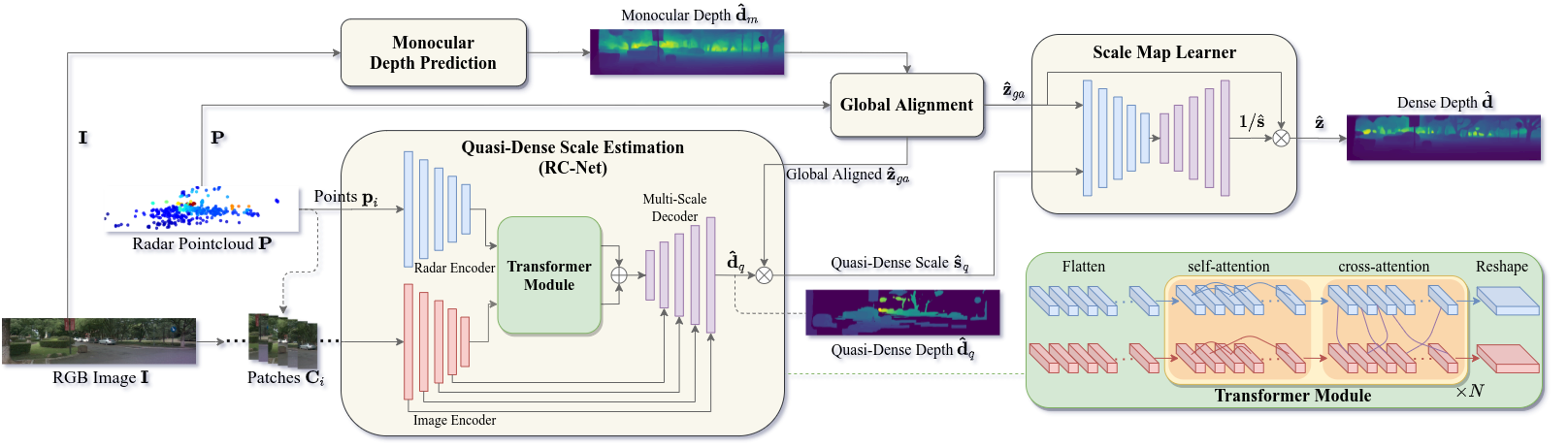}
      \vspace{-1pt}
      \caption{The overall framework of our proposed RadarCam-Depth, comprised with four stages: monocular depth prediction, global alignment of mono-depth with sparse Radar depth, learned quasi-dense scale estimation, and scale map learner for refining local scale. $\mathbf{d}$ and $\mathbf{s}$ denotes the depth and scale, while $\mathbf{z}=1/\mathbf{d}$ is the inverse depth.}
      \label{fig:overrall}
      \vspace{-15pt} 
\end{figure*}

However, sparsity, substantial noise in Radar data, and the imperfect cross-modal association between Radar points and image pixels pose challenges for dense depth estimation. Previous Radar-Camera methods treat the dense depth estimation as a depth completion problem \cite{long2021radar, singh2023depth}. In these methods, the initial step involves associating the Radar depth to the camera pixels, generating a sparse or semi-dense depth map, which is then completed by an Unet-like network with a fusion of the Radar depth and image data. 
In this paper, we propose a novel paradigm, RadarCam-Depth, which capitalizes on robust and versatile scaleless monocular depth prediction and learns to assign metric dense scales to the mono-depth with Radar data.
Our novel paradigm offers two main benefits: (i) We circumvent the direct fusion of raw data or encodings of heterogeneous Radar and camera data, thereby preventing aliasing artifacts and preserving high-fidelity fine details in dense depth estimation (see Fig.\ref{fig:fm}). (ii) Unlike learning the depth completion with a wide convergence basin, we essentially learn to complete the sparse scale obtained by aligning Radar depth with the mono-depth, which is more accessible and conducive to effective learning.

The primary contributions of this work are as follows: 
(i) We introduce the first approach that enhances the highly generalizable, scaleless mono-depth prediction with the dense metric scale intricately inferred from the noisy and sparse Radar data.
(ii) We present a novel metric dense depth estimation framework that effectively fuses heterogeneous Radar and camera data. Our framework comprises four stages: mono-depth prediction, global scale alignment of the monocular depth, Radar-Camera quasi-dense scale estimation, and scale map learner for refining the quasi-dense scale locally.
(iii) The proposed method is extensively tested on the nuScenes benchmark and our self-collected ZJU-4DRadarCam dataset. It outperforms the state-of-the-art (SOTA) techniques, substantially enhancing Radar-Camera dense depth estimation with high metric accuracy and strong generalizability.
(iv) To fertilize future research in robust depth estimation, we will release our code and high-quality ZJU-4DRadarCam dataset, including raw 4D Radar data, RGB images, and meticulously generated ground truth from LiDAR measurements.

\section{RELATED WORK}
\label{sec:related work}

\subsection{Monocular Depth Estimation}
Monocular depth estimation is a challenging task due to the inherent scale ambiguity. Many researchers have tried to address this issue by integrating it with optical flow \cite{zhao2020towards}, uncertainty estimation \cite{poggi2020uncertainty}, semantic segmentation \cite{hoyer2023improving}, instance segmentation \cite{bian2019unsupervised} and visual odometry \cite{song2023unsupervised}. Although some previous studies \cite{ranftl2020towards, ranftl2021vision, guizilini2023towards, birkl2023midas} have achieved promising results in affine-invariant scaleless depth estimation across diverse datasets, recovering the metric scale remains a significant challenge.
Some existing methods rely on inertial data to provide scale. 
To enhance the generalization, VI-SLAM \cite{sartipi2020deep} warps the input image to match the orientation prevailing in the training dataset.
CodeVIO \cite{zuo2021codevio} proposes a tightly coupled VIO system with optimizable learned dense depth. It jointly estimates VIO poses and optimizes the predicted and encoded dense depth of specific keyframes efficiently. 
Xie et al.~\cite{xie2020video} utilize a flow-to-depth layer to refine camera poses and generate depth proposals. They solve a multi-frame triangulation problem to enhance the estimation accuracy.
Recently, Wofk et al.~\cite{wofk2023monocular} introduced a framework for metric dense depth estimation from the VIO sparse depth and monocular depth prediction, which inspires our work.
They first globally align the scaleless mono-depth with the metric VIO sparse depth and then learn to refine the dense scale of the globally aligned mono-depth.

\subsection{Depth Estimation from Radar-Camera Fusion}
The fusion of Radar and camera data for metric depth estimation is an active research topic. Lin et al. \cite{lin2020depth} introduce a two-stage CNN-based pipeline that combines Radar and camera inputs to denoise Radar signals and estimate dense depth.
Long et al. \cite{long2021full} propose a Radar-2-Pixel (R2P) network that utilizes radial Doppler velocity and induced optical flow from images to associate Radar points with corresponding pixel regions, enabling the synthesis of full-velocity information. They also achieve image-guided depth completion using Radar and video data \cite{long2021radar}. 
Another approach, DORN \cite{lo2021depth} proposed by Lo et al., extends Radar points in the elevation dimension and applies deep ordinal regression network-based \cite{fu2018deep} feature fusion. 
Unlike other methods, R4dyn \cite{gasperini2021r4dyn} creatively incorporates Radar as a weakly supervised signal into a self-supervised framework and employs Radar as an additional input to enhance the robustness. However, their method primarily focuses on vehicle targets and does not fully correlate all Radar points with a larger image area, resulting in lower depth accuracy.
Recently, Singh et al. \cite{singh2023depth} present a method that relies solely on a single image frame and Radar point cloud. Their first-stage network infers the confidence scores of Radar-Pixel correspondence, generating a semi-dense depth map. They further employ a gated fusion network to control the fusion of multi-modal Radar-Camera data and predict the dense depth.
However, all the above methods directly encode and concatenate the ambiguous Radar depth and images, confusing the learning pipeline and resulting in suboptimal depth estimation.

\section{METHODOLOGY}
\label{sec:methodology}

Our goal is to recover the dense depth $\mathbf{\hat{d}} \in \mathbb{R}^{H_0 \times W_0}_+$ from a pair of RGB image $\mathbf{I} \in \mathbb{R}^{3\times H_0\times W_0}$ and Radar point cloud $\mathbf{P} = \left\{\mathbf{p}_i | \mathbf{p}_i \in \mathbb R^3, i = 0,1,2,\cdots,k-1\right\}$ in the image coordinate. $H_0$ and $W_0$ denote the height and width of the image, respectively. Either 3D or 4D Radar usually has a small field of view with ambiguous sparse data deteriorated by intensive noises.
For cross-modal fusion, it is straightforward to project Radar points onto the image plane, generating Radar depth. However, direct fusion of the inherently ambiguous and sparse Radar depth with images, achieved by concatenating their encodings or raw data, can confuse the learning pipeline~\cite{yang2019dense, van2019sparse, wong2021unsupervised, singh2023depth}, resulting in aliasing and other undesirable artifacts in the estimated depth. In this paper, we propose to get the scaleless dense depth with existing versatile monocular depth prediction networks, then learn to augment scaleless depth with accurate metric scales from Radar data.

The framework of our Radar-Camera depth estimation method consists of four stages: scaless monocular depth prediction, global alignment (GA) of mono-depth, quasi-dense scale estimation, and scale map learner (SML) for refining dense scale locally, as shown in Fig.\ref{fig:overrall}.

\subsection{Monocular Depth Prediction}
\label{MonoPred}

\begin{figure}[h]
      \centering
      \includegraphics[width=0.48\textwidth]{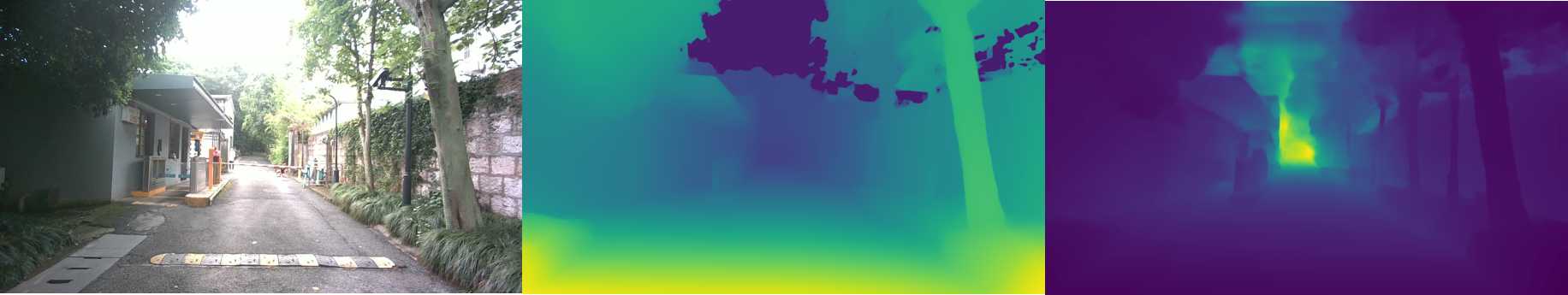}
      \captionsetup{font=small}
      \caption{Left: the input image. Middle: Mono-Pred of MiDaS v3.1 \cite{birkl2023midas}. Right: Mono-Pred of DPT-Hybrid \cite{ranftl2021vision}. Notably, MiDaS exhibits the ability to differentiate the sky.}
      \label{fig:mono}
      \vspace{-5pt} 
\end{figure}

We employ off-the-shelf networks to predict robust and accurate scaleless depth from a single-view image. 
The high quality of the mono-depth prediction furnishes a solid foundation for scale-oriented learning. In this research, we harnessed SOTA mono-depth networks, like MiDaS v3.1~\cite{ranftl2020towards, birkl2023midas} and DPT-Hybrid~\cite{ranftl2021vision} with pre-trained weights on mixed diverse datasets. Both networks are built upon transformer architecture~\cite{dosovitskiy2020image} and trained with scale and offset-invariant losses, ensuring strong generalization. They infer the relative depth relationship between pixels, producing dense depth (see Fig.\ref{fig:mono}). 
Notably, our framework is versatile and compatible with arbitrary mono-depth prediction networks that predict depth $\mathbf{\hat{d}}_m$, inverse depth $\mathbf{\hat{z}}_m$ or others.

\vspace{-5pt} 

\vspace{-5pt}
\subsection{Global Alignment}
\label{scale estimator}

We align the scaleless mono-depth prediction $\mathbf{\hat{d}}_m$ with the Radar depth originating from projecting raw Radar points $\mathbf{P}$, by a global scaling factor $\hat{s}_g$ and optional offset $\hat{t}_g$. The global aligned metric depth is calculated by $\mathbf{\hat{d}}_{ga} = \hat{s}_g \cdot \mathbf{\hat{d}}_m + \hat{t}_g$. Then, it is fed into the subsequent scale map learner (SML). There are many options for performing this global alignment between the projected Radar depth and mono-depth prediction, including:
(i) \textbf{Var}: A varying $\hat{s}_g$ for individual frame of mono-depth, calculated via root-finding algorithms \cite{forsythe1977computer, brent2013algorithms}. 
(ii) \textbf{Const}: A constant $\hat{s}_g$ for all frames of mono-depth prediction, deemed as the mean of scale estimates on the entire training samples. 
(iii) \textbf{LS}: $\hat{s}_g$ and $\hat{t}_g$ for individual frames, computed with linear least-squares optimization \cite{ranftl2020towards}. 
(iv) \textbf{RANSAC}: $\hat{s}_g$ and $\hat{t}_g$ for individual frames, computed with linear least-squares while incorporating RANSAC outlier rejection of the Radar depth.
We randomly sample 5 Radar points with valid depth values, estimate $\hat{s}_g$ and $\hat{t}_g$ with the sampled Radar depth, and adopt the first pair of $\hat{s}_g$ and $\hat{t}_g$ that yields an inlier ratio over $90\%$. The inlier is the one where the discrepancy between the Radar point depth and aligned mono-depth is under $6$m or the inverse depth discrepancy is under $0.015$.

\begin{figure*}[ht]
     \centering 
     \begin{subfigure}[t]{0.47\textwidth}
      \centering
      \tiny
    \includegraphics[width=\columnwidth]{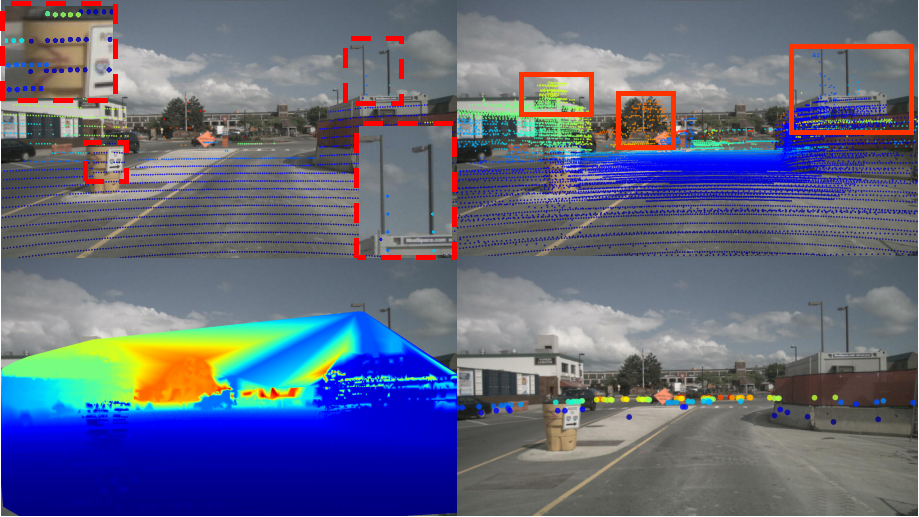}
    \captionsetup{font=scriptsize}
    \caption{}
     \AtNextBibliography{\large}{
    \label{fig:nu}
    }
     \end{subfigure}
     \hspace{20pt}
     \begin{subfigure}[t]{0.378\textwidth}
      \centering
    \includegraphics[width=\columnwidth]{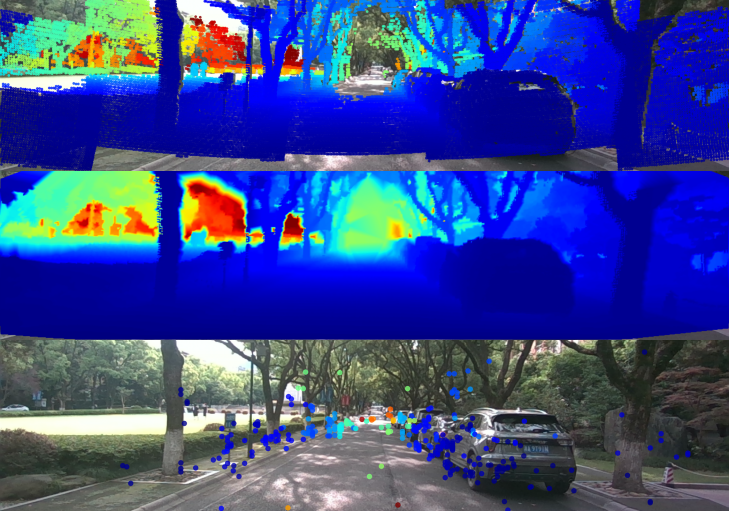}
    \captionsetup{font=scriptsize}
    \caption{}
    \label{fig:zju}
     \end{subfigure}

     \captionsetup{font=small}
     \caption{(a) Top: nuScenes dataset~\cite{caesar2020nuscenes} with LiDAR depth $\mathbf{d}_{gt}$ and accumulated LiDAR depth $\mathbf{d}_{acc}$, depth from 3D Radar point cloud $\mathbf{P}$, interpolated LiDAR $\mathbf{d}_{int}$ shown clockwise. The misalignment between LiDAR points and image pixels on this dataset is highlighted with red boxes. Depth from the 3D Radar point cloud is very sparse and non-uniformly distributed.
     (b) Our ZJU-4DRadarCam dataset with LiDAR depth $\mathbf{d}_{gt}$, interpolated LiDAR depth $\mathbf{d}_{int}$, and depth from 4D Radar point cloud $\mathbf{P}$ shown from top to bottom.
     Compared to nuScenes, the ZJU-4DRadarCam dataset offers more accurate and denser LiDAR depth and denser 4D Radar depth.}
     \label{fig:dataset compare} 
     \vspace{-10pt} 
\end{figure*}

\subsection{Quasi-Dense Scale Estimation} 
\label{RC-Net}

Due to inherent sparsity and noises in Radar data, additional enhancement of raw Radar depth is crucial before conducting the scale map learner.
To densify the sparse Radar depth obtained from projection, we exploit a transformer-based Radar-Camera data association network (shorthand RC-Net), which predicts the confidence of Radar-Pixel associations. Pixels without a direct correspondence of Radar point during projection might be associated with the depth of neighboring Radar point, thereby densifying the sparse Radar depth to a quasi-dense depth map, denoted as $\mathbf{\hat{d}}_q$.

\subsubsection{Network Architecture}
Our RC-Net (see Fig.\ref{fig:overrall}) is adapted from existing vanilla network RC-vNet \cite{singh2023depth} by further incorporating self and cross-attention \cite{sun2021loftr} in a transformer module. 
The image encoder is a standard ResNet18 backbone \cite{he2016deep} with 32, 64, 128, 128, 128 channels in each layer, and the Radar encoder is a multi-layer perceptron consisting of fully connected layers with 32, 64, 128, 128, 128 channels. The Radar features are mean pooled and reshaped to the shape of image features. Subsequently, Radar and image features are flattened and passed through $N=4$ layers of self and cross-attention, which involves a larger receptive field for the cross-modal association.
These features, combined with skip connections from intermediate layers in the encoder, are forwarded to a decoder with logit output. Finally, the logits are activated by the sigmoid function to obtain the confidence map of cross-modal associations.

\subsubsection{Confidence of Cross-Modal Associations}
For a Radar point $\mathbf{p}_i$ and a cropped image patch $\mathbf{C}_i \in \mathbb{R}^{3\times H\times W}$ in its projection vicinity, we use RC-Net $h_{\theta}$ to obtain a confidence map $\mathbf{\hat{y}}_i=h_{\theta}(\mathbf{C}_i, \mathbf{p}_i)\in [0,1]^{H\times W}$, which describes probability of whether the pixels in $\mathbf{C}_i$ corresponds to $\mathbf{p}_i$. With $k$ points in a Radar point cloud $\mathbf{P}$, the forward pass generates $k$ confidence maps for individual Radar points.
Therefore, each pixel $\mathbf{x}_{uv}$, $(u\in[0,W_0-1], v\in[0,H_0-1])$ within image $\mathbf{I}$ has $n \in [0,k]$ associated Radar point candidates. By selecting the maximum score above the threshold $\tau$, we can find the corresponding Radar point $\mathbf{p}_\mu$ for pixel $\mathbf{x}_{uv}$, and assign the depth of $\mathbf{p}_\mu$ to $\mathbf{x}_{uv}$.  Ultimately, this stage yields a quasi-dense depth map $\mathbf{\hat{d}}_q \in \mathbb{R}^{H_0 \times W_0}_ +$:
\begin{align}
    \hat{\mathbf{d}}_q(u,v) =
    \begin{cases} 
    d(\mathbf{p}_\mu),\ \ \ \  \text{if} \ \mathbf{\hat{y}}_\mu(x_{uv})  > \ \tau \\
    \text{None}, \ \ \ \ \  \text{otherwise}
    \end{cases}
\end{align}
where $\mu = \underset{i}{\arg\max}\ \mathbf{\hat{y}}_i(x_{uv})$,  and $d(\cdot)$ returns the depth value.  Finally, the quasi-dense scale map $\mathbf{\hat{s}}_q$ is calculated from $\mathbf{\hat{s}}_q=\mathbf{\hat{d}}_{q} / \mathbf{\hat{d}}_{ga}$, and its inverse $1/\mathbf{\hat{s}}_q$ is subsequently fed into the scale map learner.

\subsubsection{Training}
For the nuScenes dataset, we first project multiple frames to the current LiDAR frame $\mathbf{d}_{gt}$ to obtain the cumulative LiDAR depth $\mathbf{d}_{acc}$. After that, linear interpolation in log space \cite{barber1996quickhull} is performed on $\mathbf{d}_{acc}$ to obtain $\mathbf{d}_{int}$. Because of its density, $\mathbf{d}_{gt}$ is directly interpolated without accumulation for the ZJU-4DRadarCam dataset. 
For supervision, we use $\mathbf{d}_{int}$ to build binary classification labels $\mathbf{y}_{i} \in \left\{0,1\right\}^{H\times W}$, where points with a depth difference less than 0.5m from the Radar point are labeled as positive.
After constructing $\mathbf{y}_{i}$, we minimize the binary cross-entropy loss:
\begin{align}
  \begin{split}
    \mathcal{L}_{BCE}=\frac{1}{|\Omega|}\sum_{x\in \Omega}-(\mathbf{y}_{i}(x)\log \mathbf{\hat{y}}_i(x)\\
    +(1-\mathbf{y}_{i}(x))\log(1-\mathbf{\hat{y}}_i(x)))
    \end{split}
\end{align}
where $\Omega \subset \mathbb{R}^2$ denotes the image region of $\mathbf{C}_i$, $x \in \Omega$ is a pixel coordinate, and $\mathbf{\hat{y}}_i = h_{\theta}(\mathbf{C}_i,\mathbf{p}_i)$ is the confidence of correspondence.

\begin{figure*}[ht]
     \centering 
     \begin{subfigure}[t]{0.30\textwidth}
      \centering
    \includegraphics[width=\columnwidth]{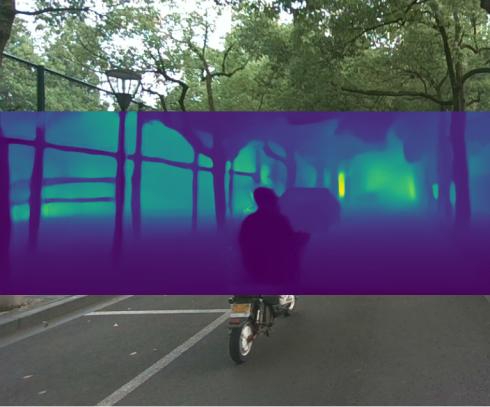}
    \captionsetup{font=scriptsize}
    \caption{}
    \label{fig:cut}
     \end{subfigure}
     \hspace{20pt}
     \begin{subfigure}[t]{0.585\textwidth}
      \centering
      \tiny
    \includegraphics[width=\columnwidth]{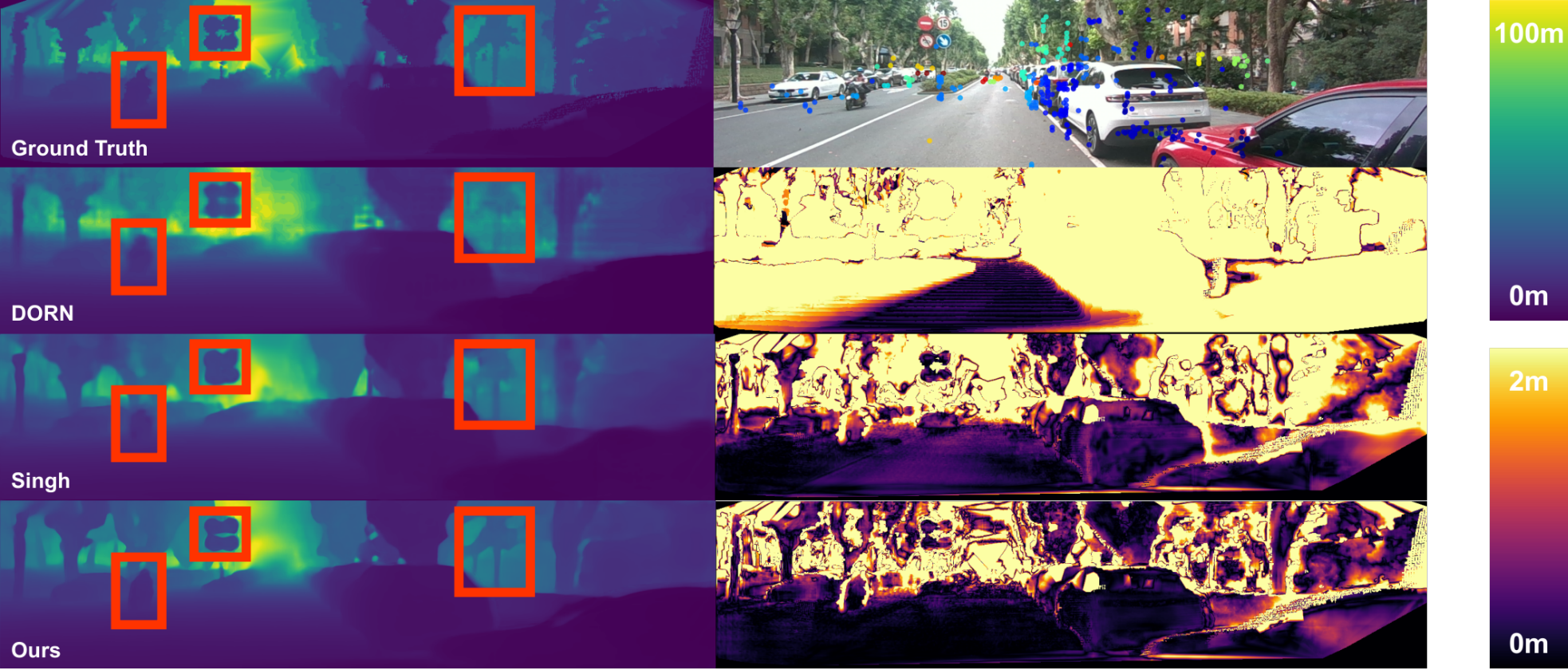}
    \captionsetup{font=scriptsize}
    \caption{}
     \AtNextBibliography{\large}{
    \label{fig:compare}
    }
     \end{subfigure}
        \vspace{-2pt}
     \captionsetup{font=small}
     \caption{
      (a) Our metric depth estimation over the input image in a large-scale scenario.
      (b) Top row shows the ground truth depth $\mathbf{d}_{int}$ and Radar points $\mathbf{P}$ projected into image $\mathbf{I}$. The rest rows from top to bottom depict the depth estimations of  \cite{lo2021depth}, \cite{singh2023depth}, and our RadarCam-Depth and the corresponding error maps. Our method demonstrates much higher accuracy and fine details.
     }
     \label{fig:compare} 
     \vspace{-10pt} 
\end{figure*}

\subsection{Scale Map Learner}
\label{SML}
\subsubsection{Network Architecture}
Inspired by  \cite{wofk2023monocular}, we construct a scale map learner (SML) network based on MiDaS-small \cite{ranftl2020towards} architecture. SML aims to learn a pixel-level dense scaling map for $\mathbf{\hat{d}}_{ga}$, which completes the quasi-dense scale map and refine the metric accuracy of $\mathbf{\hat{d}}_{ga}$. SML requires concatenated $\mathbf{\hat{z}}_{ga}$ and $1/\mathbf{\hat{s}}_q$ as input. The empty locations in $\mathbf{\hat{s}}_q$ are filled with ones. SML regresses a dense scale residual map $\mathbf{r}$, where values can be negative. We obtain the final scale map via $1/\mathbf{\hat{s}}=\text{ReLU}(1 + \mathbf{r})$, and the final metric depth estimation is computed by  $\mathbf{\hat{d}} = \mathbf{\hat{s}} /\mathbf{\hat{z}}_{ga}$.

\subsubsection{Training}
Ground truth depth $\mathbf{d}_{gt}$ is obtained from projecting 3D LiDAR points. LiDAR depth is further interpolated to get a densified depth $\mathbf{d}_{int}$. During training, we minimize the difference between the estimated metric dense depth $\hat{\mathbf{d}}$ and $\mathbf{d}_{gt}$, $\mathbf{d}_{int}$ with a smoothed L1 penalty:
\begin{align}
    \mathcal{L}_{SML}=\mathcal{L}(\mathbf{d}_{int},\hat{\mathbf{d}})+\lambda_{gt}\mathcal{L}(\mathbf{d}_{gt},\hat{\mathbf{d}})
\end{align}
\begin{align}
    \mathcal{L}(\mathbf{d},\hat{\mathbf{d}})=
    \begin{cases} 
    \frac{1}{|\Omega_d|}\underset{x\in \Omega_d}{\sum} (\mathbf{r}_d(x)-\beta/2), \  \text{if}\ \mathbf{r}_d(x)<\beta\\
    \frac{1}{|\Omega_d|}\underset{x\in \Omega_d}{\sum} (\mathbf{r}_d(x)^2/2\beta),\ \ \ \text{otherwise}
    \end{cases}
\end{align}
where $\mathbf{r}_d(x)=|\mathbf{d}(x)-\hat{\mathbf{d}}(x)|$, $\lambda_{gt}$ is the weight of $\mathcal{L}_{gt}$, $\Omega_{d} \subset \Omega$ denotes the region where ground truth has valid depth values. $\beta$ is set to 1 in our practice.

\section{EXPERIMENTS}
\label{sec:experiement}
\subsection{Datasets}
\subsubsection{NuScenes Dataset}
We first evaluate our method on nuScenes benchmark \cite{caesar2020nuscenes}. NuScenes dataset encompasses data collection across 1000 scenes in Boston and Singapore with LiDAR, 3D Radar, camera, and IMU sensors. It comprises around 40000 synchronized Radar-Camera keyframes. We followed the same data splits as \cite{singh2023depth} with 850 scenes for training and validation, and 150 for testing. The test split is officially offered by nuScenes v1.0.

\begin{figure}[h]
      \centering
      \includegraphics[width=0.48\textwidth]{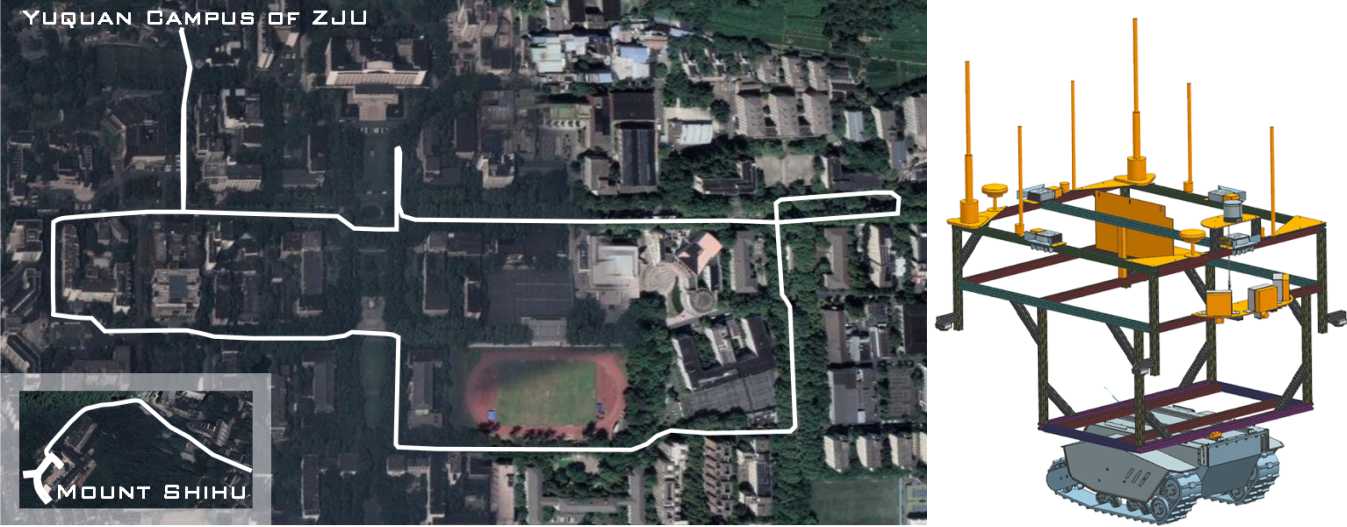}
      \captionsetup{font=small}
      \caption{Our data collection routes and the CAD model of our robot.}
      \label{fig:dataset}
      \vspace{-10pt} 
\end{figure}

\subsubsection{ZJU-4DRadarCam Dataset}
For extensive evaluation, we collected our own dataset, named ZJU-4DRadarCam, using a ground robot (Fig.\ref{fig:dataset}) equipped with Oculii's EAGLE 4D Radar, RealSense D455 camera, and RoboSense M1 LiDAR sensors.
Our dataset consists of various driving scenarios, including urban and wilderness environments. Compared to nuScenes dataset, our ZJU-4DRadarCam offers 4D Radar data with denser measurements. Besides, we provide denser LiDAR depth for supervision and evaluation (see Fig.\ref{fig:dataset compare}). Our ZJU-4DRadarCam comprises a total of 33,409 synchronized Radar-Camera keyframes, split into 29312 frames for training and validation and 4097 frames for testing.

\subsection{Training Details and Evaluation Protocol}
For the nuScenes dataset, following \cite{singh2023depth}, we accumulate the individual $\mathbf{d}_{gt}$ of 160 frames nearby to get $\mathbf{d}_{acc}$, which is then interpolated to yield $\mathbf{d}_{int}$. Dynamic objects are masked out during the above process. For the ZJU-4DRadarCam dataset, we directly interpolate $\mathbf{d}_{gt}$ to obtain $\mathbf{d}_{int}$ with linear interpolation \cite{barber1996quickhull} in the log space of depth.

For training the RC-Net on nuScenes, with an input image size of $900 \times 1600$, the size of the cropped patch during confidence map formation is set to $900\times288$. 
For the training on ZJU-4DRadarCam,  the input image size is $300 \times 1280$, while the patch size is $300\times100$. 
We employ the Adam optimizer with $\beta_1$=0.9 and $\beta_2$=0.999, and a learning rate of $2e^{-4}$ for 50 epochs. Data augmentations, including horizontal flipping, saturation, brightness, and contrast adjustments, are applied with a 0.5 probability. 
We train our RC-Net for 50 epochs with an NVIDIA RTX 3090 GPU, taking approximately 14 hours with a batch size of 6.

We adopt MiDaS-Small architecture for our SML, where the encoder backbone is initialized with pre-trained ImageNet weights \cite{deng2009imagenet}, and other layers are randomly initialized. The input data is resized and cropped to $288 \times 384$. We use an Adam optimizer with $\beta_1=0.9$ and $\beta_2=0.999$. The initial learning rate is set to  $2e^{-4}$ and reduced to $5e^{-5}$ after 20 epochs. Training SML for 40 epochs takes about 24 hours with a batch size of 24.

Some widely adopted metrics from the literature are used for evaluating the depth estimations, including mean absolute error (MAE), root mean squared error (RMSE), absolute relative error (AbsRel), squared relative error (SqRel), the errors of inverse depth (iRMSE, iMAE), and $\delta_1$~\cite{sun2021neuralrecon}. 
To better illustrate our experimental details, the demo video is available at \url{https://youtu.be/JDn0Sua5d9o}.

\begin{figure}[h]
      \centering
      \includegraphics[width=0.48\textwidth]{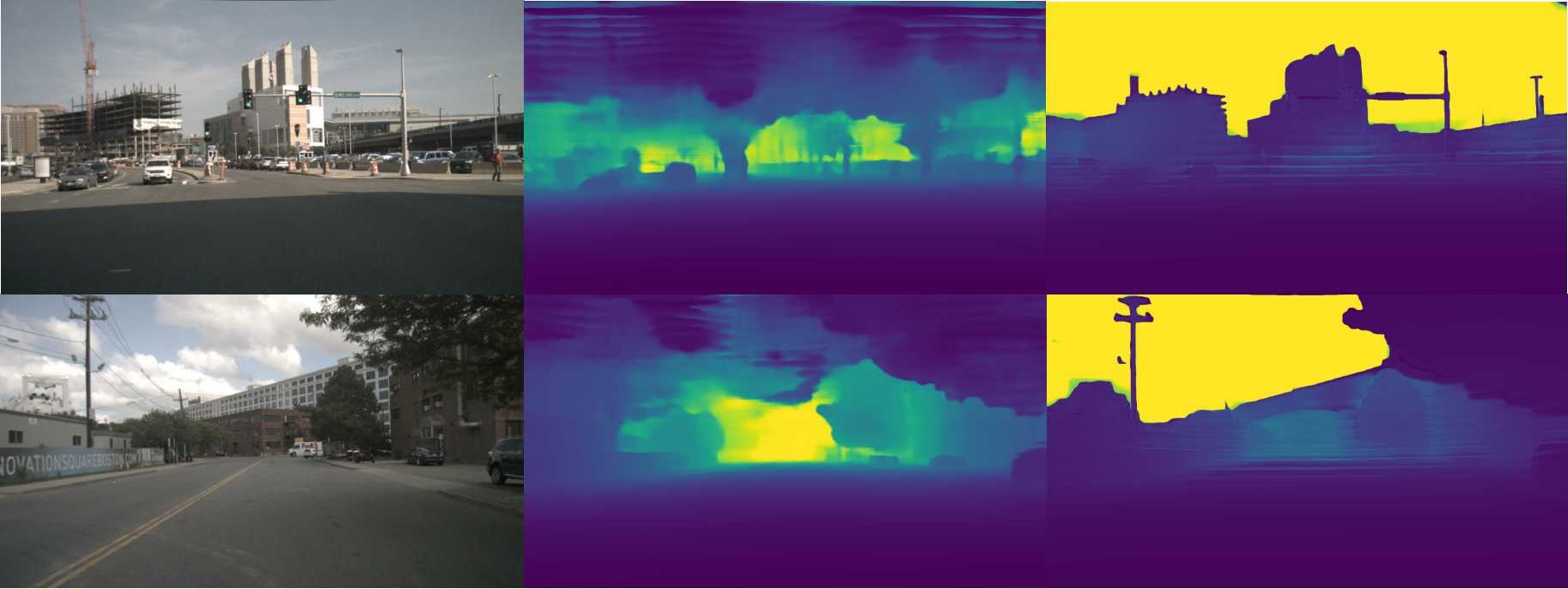}
      \captionsetup{font=small}
      \caption{From left to right are the input images, depth estimation of \cite{singh2023depth}, and depth estimation of our RadarCam-Depth on nuScenes benchmark. Ours accurately differentiates the sky region and preserves fine details on object boundaries.}
      \label{fig:nus}
      \vspace{-12pt}
\end{figure}

\subsection{Evaluation on NuScenes}
\label{evaluation on nu}
We evaluate the metric dense depth against $\mathbf{d}_{gt}$ within the range of 50, 70, and 80 meters (see Tab.\ref{nuscenes test}). Our proposed RadarCam-Depth outperforms all the compared Radar-Camera methods and surpasses the second best method~\cite{singh2023depth} by a large margin at all ranges. Specifically, we observe 25.6\%, 23.4\%, and 22.5\% reductions in MAE and 20.9\%, 20.2\%, and 19.4\% drops in RMSE for 50m, 70m, and 80m, respectively. 
We attribute the outstanding performance of RadarCam-Depth to the reasonable monocular prediction $\mathbf{\hat{d}}_{m}$ and our scale learning strategy.
Notably, RadarCam-Depth solely relies on a single-frame image and a Radar point cloud, obviating the need to aggregate multi-frame data.
The ``Radar" and ``Image" columns in Tab.\ref{nuscenes test} specify the quantities of point clouds and images used as inputs for various methods.
%
For nuScenes dataset, we adopt the sky-sensitive pre-trained model, MiDaS v3.1, as our monocular depth prediction network, which can accurately differentiate the sky from others (see Fig.\ref{fig:mono}). However, since the depth estimations in sky regions are not associated with corresponding LiDAR depth, they are not counted during metric evaluations. Some snapshots of depth estimations from different methods are shown in Fig.\ref{fig:nus}.

\begin{table}[th]
\centering
\caption{\textsc{Evaluations on NuScenes} (mm)}
\label{nuscenes test}
\resizebox{\linewidth}{!}{
\begin{tabular}{c|c|c|c|c|c}
\hline\hline
{\textbf{Eval Dist}} & \textbf{Method} & \textbf{Radar} & \textbf{Image} & \textbf{MAE} & \textbf{RMSE}\\

\hline
\multirow{5}*{50m} & RC-PDA \cite{long2021radar} & 5 & 3 & 2225.0 & 4156.5 \\
{} &  RC-PDA with HG \cite{long2021radar}& 5 & 3 & 2315.7 & 4321.6 \\
{} &  DORN \cite{lo2021depth} & 5(x3) & 1 & 1926.6 & 4124.8 \\
{} &  Singh \cite{singh2023depth} & 1 & 1  & 1727.7  & 3746.8 \\
{} &  \textbf{RadarCam-Depth}  & 1 & 1  & \textbf{1286.1}  & \textbf{2964.3} \\

\hline
\multirow{5}*{70m} & RC-PDA \cite{long2021radar} & 5 & 3 & 3326.1 & 6700.6 \\
{} &  RC-PDA with HG \cite{long2021radar} & 5 & 3 & 3485.6 & 7002.9 \\
{} &  DORN \cite{lo2021depth} & 5(x3) & 1 & 2380.6 & 5252.7 \\
{} &  Singh  \cite{singh2023depth}& 1 & 1  & 2073.2  & 4590.7 \\
{} &  \textbf{RadarCam-Depth}  & 1 & 1  & \textbf{1587.9}  & \textbf{3662.5} \\

\hline
\multirow{9}*{80m} & RC-PDA \cite{long2021radar} & 5 & 3 & 3713.6 & 7692.8 \\
{} &  RC-PDA with HG \cite{long2021radar} & 5 & 3 & 3884.3 & 8008.6 \\
{} &  DORN \cite{lo2021depth} & 5(x3) & 1 & 2467.7 & 5554.3 \\
{} &  Lin \cite{lin2020depth} & 3 & 1 & 2371.0 & 5623.0 \\
{} &  R4Dyn \cite{gasperini2021r4dyn} & 4 & 1  & N/A & 6434.0 \\
{} &  Sparse-to-dense \cite{ma2018sparse} & 3 & 1 & 2374.0 & 5628.0 \\
{} &  PnP \cite{wang2018plug} & 3 & 1 & 2496.0 & 5578.0 \\
{} &  Singh \cite{singh2023depth} & 1 & 1  & 2179.3 & 4898.7 \\
{} &  \textbf{RadarCam-Depth}  & 1 & 1  & \textbf{1689.7}  & \textbf{3948.0} \\

\hline\hline
\end{tabular}
}
\vspace{-15pt} 
\end{table}

\subsection{Evaluation on ZJU-4DRadarCam}

We follow a similar way to Sec.\ref{evaluation on nu} for the evaluations on the ZJU-4DRadarCam dataset. For the mono-depth prediction in our framework, we tried both MiDaS v3.1 \cite{birkl2023midas}, and DPT-Hybrid \cite{ranftl2021vision} models. The evaluations of the metric dense depth estimations from various methods are presented in Tab.\ref{zju test}, where our approach at different configurations of mono-depth network and global alignment options are marked in bold. After a comprehensive evaluation, we observe that our methods with the DPT model perform better for depth metrics, while the methods with MiDaS demonstrate higher accuracy for inverse depth metrics. 
Overall, our proposed methodology exhibits significant improvements compared to existing Radar-Camera methods \cite{singh2023depth} and \cite{lo2021depth} (Fig.\ref{fig:compare}). Compared to the second best \cite{singh2023depth}, the best configuration of our method shows 40.2\%, 40.1\%, and 40.2\% reductions in MAE within ranges of 50m, 70m, and 80m, respectively.

\vspace{0pt}
\begin{table}[th]
\centering
\caption{\textsc{Evaluations on ZJU-4DRadarCam} (mm)}
\label{zju test}
\resizebox{\linewidth}{!}{
\begin{tabular}{c|c|c|c|c|c|c|c|c}
\hline\hline
{\textbf{Dist}} & \textbf{Method} & \textbf{MAE} & \textbf{RMSE} & \textbf{iMAE} & \textbf{iRMSE} & \textbf{AbsRel}& \textbf{SqRel} & $\mathbf{\delta_1}$ \\

\hline
\multirow{7}*{50m} & DORN \cite{lo2021depth} & 2210.171 &  4129.691 & 19.790 & 31.853 & 0.157 & 939.348 & 0.783 \\
{} & Singh \cite{singh2023depth} &  1785.391 & 3704.636 & 18.102 & 35.342 & 0.146 & 966.133 & 0.831 \\
{} & DPT+Var+RC-vNet \cite{singh2023depth} & 1243.339 & 3045.853 & 12.111 & 24.377 & 0.098 & 644.709 & 0.896 \\

{} & \textbf{DPT+Const+RC-Net} & 1082.927 & \textbf{2803.180} & 10.885 & 23.227 & 0.089 & \textbf{561.834} & 0.920 \\
{} & \textbf{DPT+Var+RC-Net} & \textbf{1067.531} & 2817.362 & 10.508 & 22.936 & 0.087 & 575.838 & 0.922 \\

{} & \textbf{MiDaS+Var+RC-Net} & 1177.257 & 3009.135 & 10.255 & \textbf{22.385} & 0.090 & 630.222 & 0.924 \\
{} & \textbf{MiDaS+LS+RC-Net} & 1083.691 & 2868.950 & \textbf{10.059} & 22.388 & \textbf{0.086} & 588.091 & \textbf{0.928} \\

\hline
\multirow{7}*{70m} & DORN \cite{lo2021depth} & 2402.180 & 4625.231 & 19.848 & 31.877 & 0.160 & 1021.805 & 0.777\\
{} & Singh \cite{singh2023depth} & 1932.690 & 4137.143 & 17.991 & 35.166 & 0.147 & 1014.454 & 0.828 \\
{} & DPT+Var+RC-vNet \cite{singh2023depth} & 1337.649 & 3358.212 & 12.047 & 24.294 & 0.099 & 672.084 & 0.894 \\

{} & \textbf{DPT+Const+RC-Net} & 1178.046 & 3121.317 & 10.824 & 23.149 & 0.090 & \textbf{589.377} & 0.918 \\
{} & \textbf{DPT+Var+RC-Net} & \textbf{1157.014} & \textbf{3117.721} & 10.444 & 22.853 & 0.087 & 601.052 & 0.921 \\

{} & \textbf{MiDaS+Var+RC-Net} & 1280.124 & 3323.488 & 10.189 & \textbf{22.300} & 0.091 & 658.416 & 0.922 \\
{} & \textbf{MiDaS+LS+RC-Net} & 1177.253 & 3179.615 & \textbf{9.996} & 22.305 & \textbf{0.086} & 614.801 & \textbf{0.926} \\

\hline
\multirow{7}*{80m} & DORN \cite{lo2021depth} & 2447.571&  4760.016&  19.856&  31.879&  0.161&  1038.919&  0.776 \\
{} & Singh \cite{singh2023depth} & 1979.459 & 4309.314 & 17.971 & 35.133 & 0.147 & 1034.148 & 0.828 \\
{} & DPT+Var+RC-vNet \cite{singh2023depth} & 1365.383 & 3467.245 & 12.033 & 24.277 & 0.099 & 682.126 & 0.894 \\

{} & \textbf{DPT+Const+RC-Net} & 1206.541 & 3239.331 & 10.812 & 23.133 & 0.090 & \textbf{599.674} & 0.918 \\
{} & \textbf{DPT+Var+RC-Net} & \textbf{1183.471} & \textbf{3228.999} & 10.432 & 22.838 & 0.088 & 610.501 & 0.920\\

{} & \textbf{MiDaS+Var+RC-Net} & 1309.859 & 3431.046 & 10.176 & \textbf{22.282} & 0.091 & 668.038 & 0.922 \\
{} & \textbf{MiDaS+LS+RC-Net} & 1205.137 & 3295.520 & \textbf{9.984} & 22.289 & \textbf{0.086} & 624.864 & \textbf{0.926} \\

\hline\hline
\end{tabular}
}
\vspace{0pt} 
\end{table}

We report our proposed method's runtime at the DPT-based mono-depth prediction configuration. The average processing times per frame are shown in Tab.\ref{runtime}. Note that Mono-Pred and GA can run simultaneously with RC-Net. Regarding different scale global alignment methods, GA (Var) and GA (LS) exhibit relatively fast speeds, while GA (RANSAC) is significantly slow and not advocated.
\vspace{0pt}
\begin{table}[th]
\centering
\caption{\textsc{Runtime Test of our Modules} (s)}
\label{runtime}
\resizebox{0.85\linewidth}{!}{
\begin{tabular}{ccccccc}
\hline\hline
\textbf{Mono-Pred} & \textbf{GA (Const)} & \textbf{GA (Var)} & \textbf{GA (LS)} & \textbf{GA (RANSAC)} & \textbf{RC-Net}  & \textbf{SML}\\
\hline
0.0651 & - & 0.0624 & 0.0044 & 2.2903 & 0.2704 & 0.1227\\
\hline\hline
\end{tabular}
}
\vspace{-10pt} 
\end{table}

\subsection{Ablation}
\subsubsection{Transformer Module}

We commence our analysis by focusing on the transformer mechanism incorporated within our novel quasi-dense scale estimation module, RC-Net. When the transformer component is disengaged, the architecture is identical to the pre-existing vanilla network, RC-vNet \cite{singh2023depth}. Our evaluation is conducted on the ZJU-4DRadarCam dataset, and the results are presented in Tab.\ref{radarnet test}. The comparison results are the error in quasi-dense depth estimation $\mathbf{\hat{d}}_q$ against ground truth $\mathbf{d}_{gt}$ within a range of 80 meters. Our RC-Net consistently outperforms RC-vNet across all evaluation metrics, as delineated in Tab.\ref{radarnet test}.
Furthermore, as indicated in Tab.\ref{zju test}, when integrated with the DPT+Var framework, DPT+Var+RC-Net exhibits notable performance superiority over its RC-vNet counterpart.

\vspace{0pt}
\begin{table}[th]
\centering
\caption{\textsc{Ablation of Transformer Module} (mm)}
\label{radarnet test}
\resizebox{\linewidth}{!}{
\begin{tabular}{c|cccccccc}
\hline\hline
\textbf{Dataset} & \textbf{Method} & \textbf{MAE} & \textbf{RMSE} & \textbf{iMAE} & \textbf{iRMSE} & \textbf{Output Pts} \\

\hline
\multirow{2}*{ZJU-4D} & RC-vNet \cite{singh2023depth} & 1308.742 & 3339.697 & 20.418 & 38.540 & 172567.846 \\
{} & RC-Net & \textbf{1083.305} & \textbf{3052.870} & \textbf{16.203} & \textbf{33.414} & \textbf{178713.881} \\

\hline\hline
\end{tabular}
}
\vspace{-10pt} 
\end{table}

\subsubsection{Global Alignment}

Following the discussions in Sec.\ref{scale estimator}, we systematically assess the four options for globally aligning the scale of mono-depth predictions.
It is worth mentioning that we set a termination criterion of 400 iterations for the GA (RANSAC), at which point we halt the iterative process and select the values of $\hat{s}_g$ and $\hat{t}_g$ that yield the highest inlier ratio. 
The evaluation results of $\hat{\mathbf{d}}_{ga}$ are presented in Tab.\ref{ga test} within a range of 80 m. Combining the runtime performance in Tab.\ref{runtime}, the best GA method for ZJU-4D (DPT) uses variable $\hat{s}_g$ (Var), and the optimal method for ZJU-4D (MiDaS) is least-squares for $\hat{s}_g$ and $\hat{t}_g$ (LS). 
The experiments showcase that estimating $\hat{t}_g$ for DPT leads to substantial inverse errors, significantly degrading the performance of the subsequent SML  (conducted in inverse space). However, ZJU-4D (MiDaS) require simultaneous estimating $\hat{s}_g$ and $\hat{t}_g$ to achieve higher accuracy.

\vspace{-5pt}

\begin{table}[th]
\centering
\caption{\textsc{Ablation of GA Module} (mm)}
\label{ga test}
\resizebox{\linewidth}{!}{
\begin{tabular}{c|ccccccccc}
\hline\hline
\textbf{Data} & \textbf{Method} & \textbf{MAE} & \textbf{RMSE} & \textbf{iMAE} & \textbf{iRMSE} & \textbf{AbsRel} & \textbf{SqRel} & $\mathbf{\delta_1}$\\

\hline
\multirow{4}*{ZJU-4D (DPT)} & {Const} & 4733.158 & \textbf{6926.261} & 37.942 & 53.913 & 0.392 & \textbf{2501.383} & 0.343 \\
{} & {Var} & \textbf{4726.168} & 6940.025 & \textbf{36.531} & \textbf{52.569} & \textbf{0.386} & 2569.835 & \textbf{0.374} \\
{} & LS & 5671.214 & 7409.278 & 111.292 & 530.436 & 0.552 & 4069.064 & 0.271 \\
{} & RANSAC & 5963.904 & 7662.980 & 336.568 & 1294.117 & 0.614 & 4732.633 & 0.277 \\

\hline
\multirow{4}*{ZJU-4D (MiDaS)} & Const & 14008.482 & 245011.138 & 38.973 & 51.691 & 0.752 & 25697452.430 & 0.358 \\
{} & Var & 7119.828 & 14297.549 & 32.285 & 46.340 & 0.468 & 13255.805 & 0.431 \\
{} & {LS} & \textbf{4799.150} & \textbf{7968.478} & 35.670 & 51.559 & 0.390 & \textbf{4659.651} & 0.394 \\
{} & {RANSAC} & 5113.080 & 11063.605 & \textbf{23.920} & \textbf{37.881} & \textbf{0.347} & 13322.258 & \textbf{0.631} \\

\hline\hline
\end{tabular}
}
\vspace{-10pt} 
\end{table}

\section{Conclusion}
\label{sec:conclusion}
This paper presents a novel method for estimating dense metric depth by integrating monocular depth prediction with the scale from sparse and noisy Radar point clouds. We propose a dedicated four-stage framework that effectively combines the high-fidelity fine details of the image and the absolute scale of Radar data, surmounting the inherent challenges of detail loss and the imprecision of metrics that manifest in existing methods based on the direction fusion of Radar and image data or their encodings.
Our experimental findings unequivocally demonstrate a significantly superior performance of the proposed methodology over the compared baseline, as substantiated by both quantitative and qualitative assessments. In general, we introduce a pioneering metric depth estimation solution, which is rigorously validated and suitable for application on fusing cameras with 3D or 4D Radars.
In our future endeavors, we aim to enhance the applicability and effectiveness of our proposed method by leveraging vision foundation models pre-trained with abundant data.






{
\AtNextBibliography{\small}
\printbibliography
}

\end{document}